\documentclass{llncs}
\usepackage{llncsdoc}
\usepackage{amsmath,amsfonts,amssymb,latexsym,bm}
\usepackage{bm}
\usepackage[dvipdfmx]{graphicx}
\usepackage{subfig}
\usepackage{url}
\usepackage{cite}

\usepackage{algorithm, algorithmic}
\renewcommand{\algorithmicrequire}{{\bf Input:}}
\renewcommand{\algorithmicensure}{{\bf Output:}}
\title{
Alternating optimization method based on nonnegative matrix factorizations for deep neural networks
}

\author{Tetsuya Sakurai$^{1,2}$ \and Akira Imakura$^1$ \and Yuto Inoue$^1$ \and Yasunori Futamura$^1$}
\institute{Department of Computer Science, University of Tsukuba \and JST/CREST}
\begin{document}
\maketitle
\begin{abstract}
The backpropagation algorithm for calculating gradients has been widely used in computation of weights for deep neural networks (DNNs). 
This method requires derivatives of objective functions and has some difficulties finding appropriate parameters such as learning rate.
In this paper, we propose a novel approach for computing weight matrices of fully-connected DNNs by using two types of semi-nonnegative matrix factorizations (semi-NMFs). 
In this method, optimization processes are performed by calculating weight matrices alternately, and backpropagation (BP) is not used. 
We also present a method to calculate stacked autoencoder using a NMF. 
The output results of the autoencoder are used as pre-training data for DNNs. 
The experimental results show that our method using three types of NMFs attains similar error rates to the conventional DNNs with BP. 
\end{abstract}
\section{Introduction}
Deep neural networks (DNNs) attracted a great deal of attention for their high efficiency in various fields, such as speech recognition, image recognition, object detection, materials discovery. 
By using a backpropagation (BP) technique proposed by Rumelhart, et al. \cite{Rumelhart1986}, computational performance is improved for training multilayer neural networks. 
However, learning often takes a long time to converge, and it may fall into a local minimum. 
Bengio, et al. \cite{Bengio2006} proposed a method to improve general performance by pre-training with an autoencoder. 
Moreover, selection of appropriate learning rates \cite{LeCun1998} and restriction of weights as dropout \cite{Srivastava2014} have also used to minimize the expected error. 
Hinton, et al. discussed initialization of weights in \cite{Hinton2012}. 
\par
Neural networks have variations such as fully-connected networks, convolutional networks and recurrent networks. 
LeCun, et al. \cite{LeCun1998} showed that convolutional neural networks attain high efficiency for image recognition. 
In DNNs, activation functions are used to attain nonlinear properties. 
Recently, the rectified linear function (ReLU) \cite{Nair2010, Glorot2011} has often been used. 
%
\par
Feedforward neural networks are computed by multiplying weight matrices and input matrices. 
Thus, the main computations are matrix--matrix multiplications (GEMM), and accelerators such as GPUs are employed to obtain high performance \cite{Ciresan2011}. 
However, a large computational cost of neural networks is still a problem.  
%
%
\par
In this paper, we propose a novel computing method for fully-connected DNNs that uses two types of semi-nonnegative matrix factorizations (semi-NMFs).
In this method, optimization processes are performed by calculating weight matrices alternately, and BP is not used.
We also present a method to calculate a stacked autoencoder using a NMF \cite{Paatero1994, Lee1999}. 
The output results of the autoencoder are used as pre-training data for DNNs. 
\par
In the presented method, computations are represented by matrix--matrix computations, and accelerators such as GPUs and MICs can be employed like in BP computations. 
In BP computations, mini-batches are used to avoid stagnations of the optimization precess.
The use of small mini-batch sizes decreases matrix sizes and gains reductions in computations. 
The presented method also uses partitioned matrices; however, the matrix size is larger than that of conventional BP, and we expect high performance. 
\par
This paper is organized as follows. 
In Section 2, we review the conventional method of computing DNNs. 
In Section 3, we present a method for computing weights in DNNs using two types of semi-NMFs. 
We also present a method to calculate a stacked autoencoder using NMF. 
In Section 4, we show some experimental results of our proposed approach. 
Section 5 presents our conclusions.
\par
We use MATLAB colon notations throughout.
Moreover, let $A = \{a_{ij} \} \in \mathbb{R}^{m \times n}$, then $A \geq 0$ denotes that all entries are nonnegative: $a_{ij} \geq 0$.
\section{Computation of deep neural networks}
Let $n_{\rm in}, n_{\rm out}, m$ be sizes of input and output units and the training data, respectively.
Moreover, let $X \in \mathbb{R}^{n_{\rm in} \times m}$ and $Y \in \mathbb{R}^{n_{\rm out} \times m}$ be input and output data.
Using a weight matrix $W \in \mathbb{R}^{n_{\rm out} \times n_{\rm in}}$ and a bias vector ${\bm b} \in \mathbb{R}^{n_{\rm out}}$, the objective function of one layer of neural networks can be written as
\begin{equation*}
        E(W, {\bm b}, X, Y) = D \left( Y, f(W X + {\bm b} {\bm e}^{\rm T}) \right) + h(W, {\bm b}),
\end{equation*}
where $D(\cdot, \cdot)$ is a divergence function, ${\bm e} = [1,1, \dots, 1]^{\rm T} \in \mathbb{R}^m$, $f(U)$ is an activation function and $h(W, {\bm b})$ is a regularization term.
There are several activation functions such as sigmoid functions like the logistic function and the hyperbolic tangent function.
Recently, rectified linear unit (ReLU) has been widely used.
%
%
\par
The objective function of DNNs with $d-1$ hidden units of size $n_i, i = 1, 2, \dots, d-1$ is written as 
\begin{align}
        &E(W_1, \dots, W_d, {\bm b}_1, \dots, {\bm b}_d, X, Y) \nonumber \\
        &\quad = D \left( Y, W_d f(W_{d-1}  \cdots f(W_1 X + {\bm b}_1{\bm e}^{\rm T}) \cdots +{\bm b}_{d-1}{\bm e}^{\rm T}) + {\bm b}_d {\bm e}^{\rm T} \right) \nonumber \\
        &\phantom{\quad = } \quad + h(W_1, \dots, W_d, {\bm b}_1, \dots, {\bm b}_d).
        \label{eq:obj_dnn}
\end{align}
Here, $W_i \in \mathbb{R}^{n_{i} \times n_{i-1}}, {\bm b}_i \in \mathbb{R}^{n_{i-1}}$, $n_0 = n_{\rm in}, n_d = n_{\rm out}$.
BP algorithms, which are based on the gradient descent method using derivatives, are one of the most standard algorithms used to minimize the objective function.
\section{An alternating optimization method based on nonnegative matrix factorization}
In this paper, we consider solving the following minimization problem
\begin{equation}
        \min_{W_1, \dots, W_d} E(W_1, \dots, W_d, X, Y),
        \label{eq:min}
\end{equation}
where the objective function simplifies the objective function \eqref{eq:obj_dnn} using the square error of DNNs and is defined by
\begin{equation}
        E(W_1, \dots, W_d, X, Y) := \frac{1}{2} \| Y - W_d f(W_{d-1} \cdots f(W_1 X)\cdots ) \|_{\rm F}^2,
        \label{eq:obj}
\end{equation}
where $\| \cdot \|_{\rm F}$ is the Frobenius norm.
Here, the activation function $f(U)$ is set as ReLU.
\par
The basic concept of our algorithm to solve \eqref{eq:min} is an alternating optimization that (approximately) optimizes each weight matrix $W_i$ for $i = d, d-1, \dots, 1$, one by one.
Let $W_1^{(0)}, W_2^{(0)}, \dots, W_d^{(0)}$ be initial guesses of $W_1, \dots, W_d$, respectively.
An autoencoder to set the initial guesses will be discussed in Section 4.
In each iteration $k$, we also define objective functions
\begin{equation*}
        E_i^{(k)}(W_i,X,Y) := E(W_1^{(k-1)}, \dots, W_{i-1}^{(k-1)}, W_i, W_{i+1}^{(k)}, \dots, W_{d}^{(k)}, X, Y),
\end{equation*}
as for the $i$-th weight matrix $W_i$.
Then, we (approximately) solve the minimization problems
\begin{equation*}
        W_i^{(k)} = \arg \min_{W_i} E_i^{(k)}(W_i,X,Y)
\end{equation*}
for $i = d, d-1, \dots, 1$.
The basic concept of our proposed method is shown in Algorithm~\ref{alg:proposed_strategy}.
\begin{algorithm}[t]
\caption{The basic concept of the proposed method}
\label{alg:proposed_strategy}
\begin{algorithmic}[1]
        \STATE Set initial guesses $W_1^{(0)}, W_2^{(0)}, \dots, W_d^{(0)}$
        \FOR{$k = 1, 2, \dots$}
        \FOR{$i = d, d-1, \dots, 1$}
        \STATE Minimize (approx.) $E_i^{(k)}(W_i,X,Y)$ for $W_i$ with an initial guess $W_i^{(k-1)}$,\\
        and get $W_i^{(k)}$
        \ENDFOR
        \ENDFOR
\end{algorithmic}
\end{algorithm}
\par
Let matrices $Z_i^{(k)} \in \mathbb{R}^{n_i \times m}$ be defined as
\begin{align*}
        &Z_0^{(k)} := X, \\
        &Z_i^{(k)} := f( W_i^{(k)} f(W_{i-1}^{(k)} \cdots f(W_1^{(k)} X) \cdots )), \quad i = 1, 2, \dots, d-1.
\end{align*}
Then, in what follows, we derive our alternating optimization algorithm using semi-NMF \cite{Ding2010} and nonlinear semi-NMF.
\subsection{Optimization for $W_d$ using semi-NMF}
Using the matrix $Z_{d-1}^{(k-1)}$, the objective function for the weight matrix $W_d$ is rewritten as
\begin{equation*}
        E_d^{(k)}(W_d,X,Y) = \frac{1}{2} \| Y - W_d Z_{d-1}^{(k-1)} \|_{\rm F}^2.
\end{equation*}
Here, we note that $Z_{d-1}^{(k-1)} \geq 0$ from the definition of $Z_i^{(k-1)}$.
Therefore, we can obtain $W_d^{(k)}$ and $\widehat{Z}_{d-1}^{(k)}$ by (approximately) solving nonnegative constraint minimization problem of the form
\begin{equation}
        [W_d^{(k)}, \widehat{Z}_{d-1}^{(k)}] =  \arg \min_{W_d, Z_{d-1}} \| Y - W_d Z_{d-1} \|_{\rm F}, \quad \mbox{s.t. } Z_{d-1}  \geq  0,
        \label{eq:semi-nmf}
\end{equation}
using initial guesses $W_d^{(k-1)}, Z_{d-1}^{(k-1)}$.
This minimization problem is known as semi-NMF.
\subsection{Optimization for $W_i, i = d-1, \dots, 1$ using nonlinear semi-NMF}
From the definition of $Z_i^{(k-1)}$, we expect 
\begin{equation}
        \widehat{Z}_{d-1}^{(k)} \approx f(W_{d-1} Z_{d-2})
        \label{eq:nonlin_semi-nmf}
\end{equation}
to minimize the objective function \eqref{eq:obj}.
Then, we consider (approximately) solving the minimization problem
\begin{equation}
        [W_i^{(k)}, \widehat{Z}_{i-1}^{(k)}] = \arg \min_{W_i, Z_{i-1}} \| \widehat{Z}_{i}^{(k)} - f(W_i Z_{i-1}) \|_{\rm F}, \quad \mbox{s.t. } Z_{i-1}  \geq  0
        \label{eq:nonlinear}
\end{equation}
for $W_i$ with $i = d-1, d-2, \dots, 1$.
This minimization problem \eqref{eq:nonlinear} is a nonnegative constraint minimization problem like \eqref{eq:semi-nmf}.
However, \eqref{eq:nonlinear} has a nonlinear activation function.
In this paper, we call this problem nonlinear semi-NMF.
\par
In order to solve this nonlinear semi-NMF, we introduce an alternating minimization algorithm that minimizes nonlinear least squares problems
\begin{equation}
        \min_{W_i} \| \widehat{Z}_i^{(k)} - f(W_i Z_{i-1}^{(k-1)}) \|_{\rm F}
        \label{eq:nonlin_lsq}
\end{equation}
and
\begin{equation}
        \min_{Z_{i-1}  \geq  0} \| \widehat{Z}_i^{(k)} - f(W_i^{(k)} Z_{i-1}) \|_{\rm F},
        \label{eq:nonlin_nnlsq}
\end{equation}
one by one.
Here, we note that \eqref{eq:nonlin_nnlsq} has a nonnegative constraint on $Z_i$.
We also note that, for $i=1$, we do not require a solution of \eqref{eq:nonlin_nnlsq}, because $Z_0 = X$.
The nonlinear least squares problems \eqref{eq:nonlin_lsq} and \eqref{eq:nonlin_nnlsq} are solved by stationary iteration-like methods as shown in Algorithms \ref{alg:nonlin_lsq} and \ref{alg:nonlin_nnlsq}, where $A^\dagger$ is a pseudo-inverse of $A$.
In practice, the pseudo-inverse of $A$ is approximated using a low-rank approximation of $A$.
\begin{algorithm}[t]
\caption{An iteration method for solving nonlinear least squares $\min_{X} \| B - f(XA) \|_{\rm F}$}
\label{alg:nonlin_lsq}
\begin{algorithmic}[1]
        \STATE Set initial guess $X_0$ and parameter $\omega$
        \FOR{$s = 0, 1, \dots$}
        \STATE $R_s = B - f( X_s A)$
        \STATE $X_{s+1} = X_s + \omega R_s A^\dagger$
        \ENDFOR
\end{algorithmic}
\end{algorithm}
\begin{algorithm}[t]
\caption{An iteration method for solving nonnegative constrain nonlinear least squares $\min_{X \geq 0} \| B - f(AX) \|_{\rm F}$}
\label{alg:nonlin_nnlsq}
\begin{algorithmic}[1]
        \STATE Set initial guess $X_0$ and parameter $\omega$
        \FOR{$s = 0, 1, \dots$}
        \STATE $R_s = B - f( A X_s)$
        \STATE $X_{s+1} = f(X_s + \omega A^\dagger R_s)$
        \ENDFOR
\end{algorithmic}
\end{algorithm}
\subsection{An alternating optimization method}
Using semi-NMF \eqref{eq:semi-nmf} and nonlinear semi-NMF \eqref{eq:nonlin_semi-nmf}, the algorithm of the proposed method is summarized in Algorithm~\ref{alg:proposed}.
In practice, the input data $X$ is approximated using a low-rank approximation based on the singular value decomposition:
\begin{equation*}
        X = [U_1, U_2] \left[
                \begin{array}{cc}
                        \Sigma_1 & \\
                        & \Sigma_2
                \end{array}
        \right] \left[
                \begin{array}{c}
                        V_1^{\rm T} \\
                        V_2^{\rm T}
                \end{array}
        \right] \approx U_1 \Sigma_1 V_1^{\rm T}.
\end{equation*}
Here, we assume that all hidden units have almost the same size: $n \approx n_i$, then the computational cost of the proposed method is $O(m n^2 + d n^3)$.
%
%
\par
The proposed method can also use the mini-batch technique.
Let $X_\ell :=$ $X(:,\mathcal{J}_\ell)$ be a submatrix of the input data $X$ corresponding to each mini-batch, where $\mathcal{J}_\ell$ is the index set in the mini-batch.
Then, in order to use the mini-batch technique for the proposed method, we need to compute the low-rank approximation of $X_\ell \approx U_{\ell,1} \Sigma_{\ell,1} V_{\ell,1}^{\rm T}$, in each iteration.
We can reduce the required computational cost by reusing the results of the low-rank approximation of $X$ as follows:
\begin{equation*}
        X(:,\mathcal{J}_\ell) \approx U_{\ell,1} \Sigma_{\ell,1} V_{\ell,1}^{\rm T} \approx U_1 \Sigma_1 V_1(\mathcal{J}_\ell, :)^{\rm T}.
\end{equation*}
\par
Other improvement techniques used for BP are also expected to improve the performance of the proposed method.
\begin{algorithm}[t]
\caption{A proposed method}
\label{alg:proposed}
\begin{algorithmic}[1]
        \STATE Set initial guess $W_1^{(0)}, W_2^{(0)}, \dots, W_d^{(0)}$
        \FOR{$k = 1, 2, \dots$}
        \STATE Solve (approx.) semi-NMF \eqref{eq:semi-nmf}\\
        with initial guesses $W_d^{(k-1)}, Z_{d-1}^{(k-1)}$ and get $W_d^{(k)}, \widehat{Z}_{d-1}^{(k)}$
        \FOR{$i = d-1, \dots, 2$}
        \STATE Solve (approx.) nonlinear LSQ \eqref{eq:nonlin_lsq} by Algorithm~\ref{alg:nonlin_lsq}\\
        with an initial guess $W_i^{(k-1)}$, and get $W_i^{(k)}$
        \STATE Solve (approx.) nonnegative constrain nonlinear LSQ \eqref{eq:nonlin_nnlsq} by Algorithm~\ref{alg:nonlin_nnlsq}\\
        with an initial guess $Z_{i-1}^{(k-1)}$, and get $\widehat{Z}_{i-1}^{(k)}$
        \ENDFOR
        \STATE Solve (approx.) nonlinear LSQ \eqref{eq:nonlin_lsq} for $i = 1$ by Algorithm~\ref{alg:nonlin_lsq}\\
        with an initial guess $W_1^{(k-1)}$, and get $W_1^{(k)}$
        \STATE Set $Z_i^{(k)}$ for $i = 1, 2, \dots, d-1$
        \ENDFOR
\end{algorithmic}
\end{algorithm}
\section{An alternating optimization-based stacked autoencoder using NMF}
\begin{algorithm}[t]
\caption{A proposed stacked autoencoder}
\label{alg:proposed_ae}
\begin{algorithmic}[1]
        \FOR{$i = 1, 2, \dots, d-1$}
        \STATE Set initial guess $\widetilde{W}_i^{(0)}, {Z}_i^{(0)}$
        \FOR{$k = 1, 2, \dots, {\rm iter}_{\rm max}$}
        \STATE Solve (approx.) NMF \eqref{eq:nmf} with initial guesses $\widetilde{W}_i^{(k-1)}, {Z}_i^{(k-1)}$, and get $\widetilde{W}_i^{(k)}, \widehat{Z}_i^{(k)}$
        \STATE Solve (approx.) nonlinear LSQ $\min_{W_i} \| \widehat{Z}_i^{(k)} -f(W_i Z_{i-1}) \|_{\rm F}$ by Algorithm~\ref{alg:nonlin_lsq}\\
        with initial guess $W_i^{(k-1)}$, and get $W_i^{(k)}$
        \STATE Set $Z_{i}^{(k)} = f(W_i^{(k)} Z_{i-1})$ 
        \ENDFOR
        \ENDFOR
\end{algorithmic}
\end{algorithm}
In this section, we propose an alternating optimization-based stacked autoencoder using NMF for computing initial guesses $W_i^{(0)}$ of the proposed method (Algorithm~\ref{alg:proposed}).
Let $\widehat{Z}_0 = X$.
Then, for the stacked autoencoder, we compute the initial guesses $W_i^{(0)}$ by (approximately) minimizing
\begin{equation*}
        \min_{W_i, \widetilde{W}_i} \| \widehat{Z}_{i-1} - \widetilde{W}_i f (W_i Z_{i-1}) \|_{\rm F}
\end{equation*}
for $i = 1, 2, \dots, d-1$ like as for the DNNs.
Each minimization problem is solved by NMF as shown below
\begin{equation}
        [\widetilde{W}_i, \widehat{Z}_i] = \arg \min_{\widetilde{W}_i, Z_i  \geq  0 } \| \widehat{Z}_{i-1} - \widetilde{W}_i Z_i \|_{\rm F}
        \label{eq:nmf}
\end{equation}
and Algorithm~\ref{alg:nonlin_lsq} is used to solve the  nonlinear least squares problem as
\begin{equation}
        W_i = \min_{W_i} \| \widehat{Z}_{i} - f(W_i Z_{i-1}) \|_{\rm F}.
        \label{eq:nonlin_lsq2}
\end{equation}
By solving \eqref{eq:nmf} and \eqref{eq:nonlin_lsq2} alternatily, the algorithm for the proposed stacked autoencoder is summarized by Algorithm~\ref{alg:proposed_ae}.
\section{Performance evaluations}
In this section, we evaluate the performance of the proposed method (Algorithm~\ref{alg:proposed}) using the stacked autoencoder (Algorithm~\ref{alg:proposed_ae}) for fully-connected DNNs for MNIST \cite{LeCun1989} and CIFAR10 \cite{Krizhevsky2009}.
There are several techniques for improving the performance of BP such as affine/elastic distortions and denoising autoencoder.
These techniques are also expected to improve the performance of our algorithm.
Therefore, in this section, we just make a comparison with a simple BP.
\par
For the proposed method, the number of iterations of the autoencoder and the LSQs and the threshold of the low-rank approximation of the input data $X$ were set as $(5,10,4.0 \times 10^{-2})$ for MNIST and $(20,25,5.0 \times 10^{-3})$ for CIFAR10, respectively.
The size of the mini-batches was set as 5000 and the autoencoder was computed using only 5000 random samples.
For optimizing parameters of BP, we used ADAM optimizer \cite{Kingma2015}.
For ADAM optimizer, initial learning rates for the stacked autoencoder and for the fine tuning were set as $(10^{-3},10^{-3})$ for MNIST and $(5.0 \times 10^{-4}, 10^{-3})$ for CIFAR10, respectively.
Other parameters $\beta_1$, $\beta_2$ and $\varepsilon$ of ADAM were set as the default parameters of TensorFlow.
We used the normalized initialization \cite{Glorot2010} for initial guesses of the stacked autoencoder.
The size of the mini-batches was set as 100 and the autoencoder was computed using only 5000 random samples.
\par
The performance evaluations were carried out using double precision arithmetic on Intel(R) Xeon(R) CPU E5-2667 v3 (3.20GHz).
The proposed method was implemented in MATLAB and the BP was implemented using TensorFlow \cite{TensorFlow}.
\begin{figure}[t]
\begin{center}
\subfloat[MNIST.]{
\includegraphics[bb = 50 50 412 292, scale=0.45]{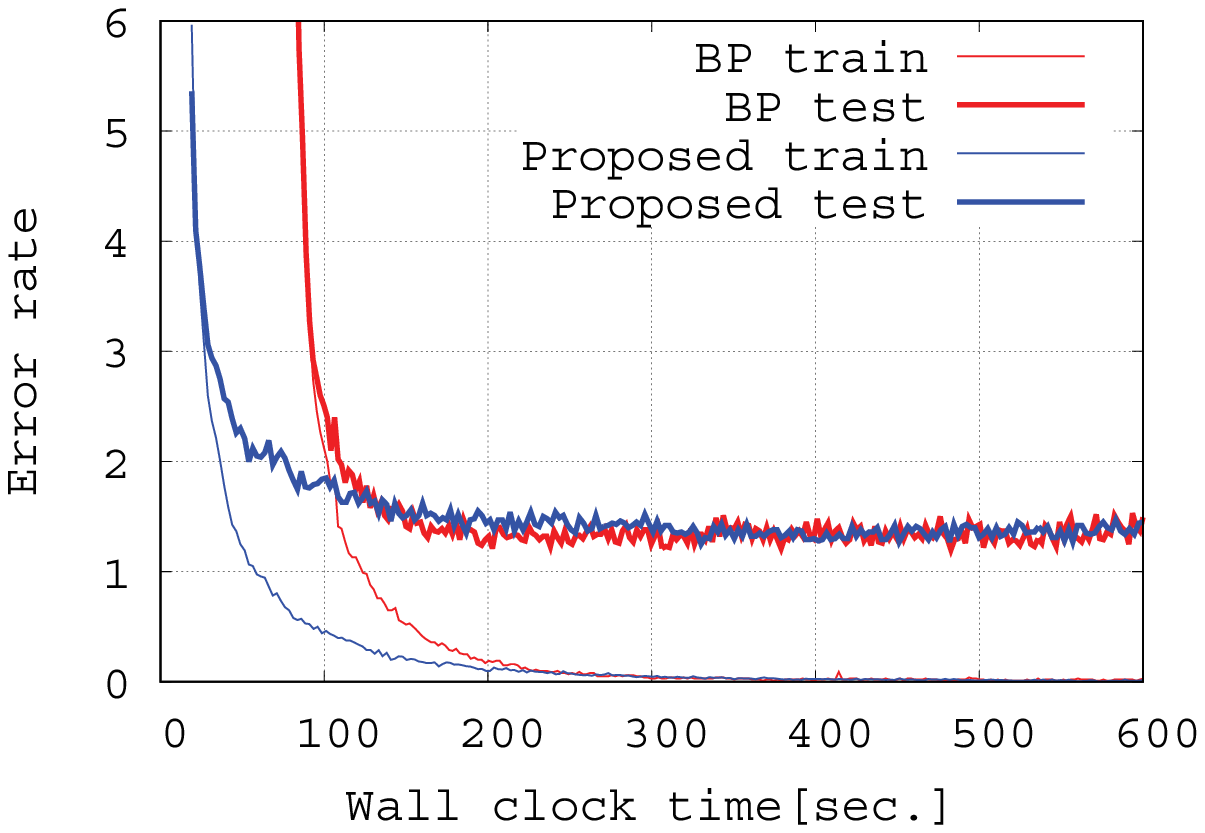}
}
\subfloat[CIFAR10.]{
\includegraphics[bb = 50 50 412 292, scale=0.45]{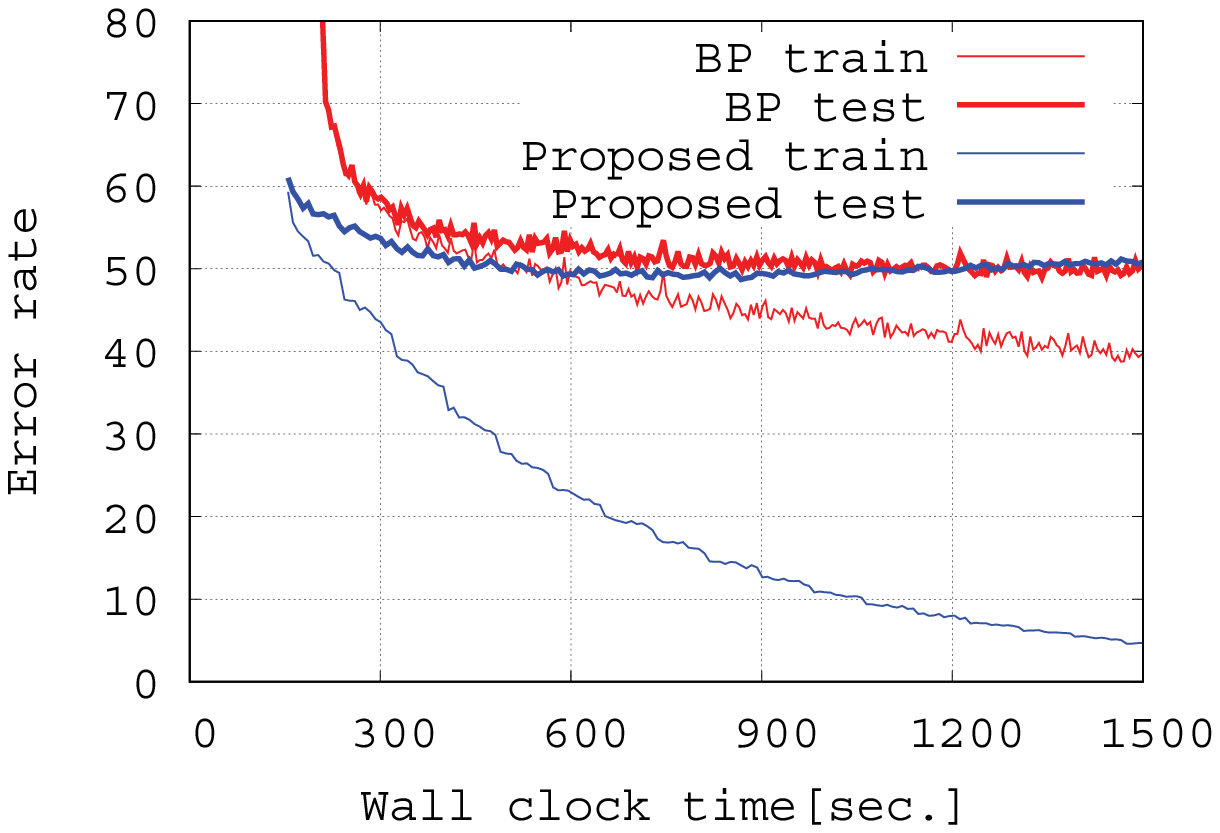}
}
\caption{Convergence history of the proposed method and BP with [1000-500] hidden units for MNIST and CIFAR10.}
\label{fig:conv}
\end{center}
\end{figure}
\begin{table}[t]
\caption{The 95\% confidence interval of the error rate and the computation time of 10 epoch of the proposed method (MATLAB) and BP (TensorFlow) with several hidden units for MNIST.}
\label{table:results}
\begin{center}
\begin{tabular}{ccccccccccrcccccccccr} \hline
 & \multicolumn{10}{c}{BP} &  & \multicolumn{9}{c}{Proposed} \\\cline{3-11}\cline{13-21}
hidden units &  & \multicolumn{7}{c}{Error rate [\%]} &  & \multicolumn{1}{c}{time} &  & \multicolumn{7}{c}{Error rate [\%]} &  & \multicolumn{1}{c}{time} \\\cline{3-9}\cline{13-19}
 &  & \multicolumn{3}{c}{train data} &  & \multicolumn{3}{c}{test data} &  & \multicolumn{1}{c}{[sec.]} &  & \multicolumn{3}{c}{train data} &  & \multicolumn{3}{c}{test data} &  & \multicolumn{1}{c}{[sec.]} \\\hline
500 &  & 0.30  & $\pm$ & 0.014  &  & 1.77  & $\pm$ & 0.03  &  & 153  &  & 3.68  & $\pm$ & 0.043  &  & 3.73  & $\pm$ & 0.09  &  & 99  \\
1000-500 &  & 0.06  & $\pm$ & 0.007  &  & 1.39  & $\pm$ & 0.08  &  & 330  &  & 0.04  & $\pm$ & 0.004  &  & 1.50  & $\pm$ & 0.06  &  & 310  \\
1500-1000-500 &  & 0.32  & $\pm$ & 0.086  &  & 1.90  & $\pm$ & 0.15  &  & 739  &  & 0.01  & $\pm$ & 0.004  &  & 1.35  & $\pm$ & 0.03  &  & 737  \\
2000-1500-1000-500 &  & 0.48  & $\pm$ & 0.132  &  & 1.84  & $\pm$ & 0.18  &  & 1589  &  & 0.00  & $\pm$ & 0.001  &  & 1.29  & $\pm$ & 0.04  &  & 1581  \\\hline
\end{tabular}
\end{center}
\end{table}
\par
Fig.~\ref{fig:conv} shows the convergence history of the proposed method and BP with [1000-500] hidden units for MNIST and CIFAR10.
Table~\ref{table:results} shows the 95\% confidence interval of the error rate and the computation time of 10 epoch of both methods with several hidden units for MNIST.
\par
These experimental results show that our method attains a similar error rate, for several hidden units, as conventional DNNs with BP.
Specifically, the proposed method achieves better error rates with deeper hidden units.
Moreover, the proposed method needs a smaller computation time for the stacked autoencoder and almost the same computation time for the fine tuning.
%
%
\section{Conclusions}
In this paper, we proposed an alternating optimization algorithm for computing weight matrices of fully-connected DNNs by using the semi-NMF and the nonlinear semi-NMF. 
We also presented a method to calculate a stacked autoencoder by using NMF. 
The experimental results showed that our method using NMF attains a similar error rate and a similar computation time to conventional DNNs with BP.
Almost the all computations of the proposed method are represented by matrix--matrix computations, and accelerators such as GPUs and MICs are employed like in BP computations. 
The proposed method also uses mini-batch technique; however, the matrix size is larger than that of conventional BP.
Therefore, we expect that the proposed method achieves high performance on recent computational environments.
\par
For future work, we will consider a bias vector, sparse regularizations and other activation functions.
Moreover, we will extend our algorithm to convolutional neural networks.
We will also consider parallel computation implementation and evaluate the performance in recent parallel environments.

\end{document}